%% file: main.tex
\documentclass[letterpaper, 10 pt, conference]{ieeeconf}  
\IEEEoverridecommandlockouts 
\usepackage{graphics}
\usepackage{lipsum}
\usepackage{amsmath}
\usepackage{amssymb}
\usepackage{cleveref}
\usepackage{adjustbox}
\usepackage{float}
\usepackage{subfigure}
\usepackage{tabulary}
\usepackage{xcolor}
\usepackage{authblk}
\usepackage{url}
\usepackage[T1]{fontenc}

\crefname{figure}{Fig}{Figs}
\crefname{equation}{eq.}{eqs.}

\title{ \LARGE \bf Imitation Learning for Autonomous Trajectory Learning of Robot Arms in Space}

\author{%
RB Ashith Shyam$^1$\thanks{University of Surrey, \texttt{a.rajendrababu@surrey.ac.uk}}%
\and
Zhou Hao$^1$%
\and
Umberto Montanaro$^1$%
\and
Gerhard Neumann$^2$\thanks{Karlshule Institute of Technology}
}

\begin{document}
\maketitle
\thispagestyle{empty}
\pagestyle{empty}


\begin{abstract}
\input{src/abstract}
\end{abstract}

\section{INTRODUCTION}
\label{sec:introduction}
\input{src/introduction.tex}
\section{RELATED WORK}
\label{sec:related_work}
\input{src/related_work.tex}

\section{DYNAMIC FORMULATION}
\label{sec:dynamic_formulation}
\input{src/dynamic_formulation.tex}

\section{DATA GENERATION FOR TRAJECTORY LEARNING}
\label{sec:trajectory_learning}
\input{src/data_generation_mpc.tex}

\section{TRAJECTORY ENCODING \& REPRODUCTION}
\label{sec:traj_encoding}
\input{src/trajectory_encoding.tex}

\section{SIMULATION RESULTS}
\label{sec:simulation}
\input{src/simulation.tex}


\section{CONCLUSION}
\label{sec:conclusion}
\input{src/conclusion.tex}


\addtolength{\textheight}{-3cm}

\section*{ACKNOWLEDGEMENT}
 This work has received funding from UKRI under the consortium named as FAIRSPACE \url{(www.fairspacehub.org)}
 
\input{src/appendices.tex}

\bibliographystyle{IEEEtran}
\bibliography{refs}
\end{document}

%% file: src/abstract.tex
{\color{black} This work adds on to the on-going efforts to provide more autonomy to space robots. Here the concept of programming by demonstration or imitation learning is used for trajectory planning of manipulators mounted on small spacecraft. For greater autonomy in future space missions and minimal human intervention through ground control, a robot arm having 7-Degrees of Freedom (DoF) is envisaged for carrying out multiple tasks like debris removal, on-orbit servicing and assembly. Since actual hardware implementation of microgravity environment is extremely expensive, the demonstration data for trajectory learning is generated using a model predictive controller (MPC) in a physics based simulator. The data is then encoded compactly by Probabilistic Movement Primitives (ProMPs). This offline trajectory learning allows faster reproductions and also avoids any computationally expensive optimizations after deployment in a space environment. It is shown that the probabilistic distribution can be used to generate trajectories to previously unseen situations by conditioning the distribution. The motion of the robot (or manipulator) arm induces reaction forces on the spacecraft and hence its attitude changes prompting the Attitude Determination and Control System (ADCS) to take large corrective action that drains energy out of the system. By having a robot arm with redundant DoF helps in finding several possible trajectories from the same start to the same target. This allows the ProMP trajectory generator to sample out the trajectory which is obstacle free as well as having minimal attitudinal disturbances thereby reducing the load on ADCS.

\textbf{Keywords}: motion planning, probabilistic movement primitives, robot manipulation, learning from demonstrations}

%% file: src/introduction.tex
{\color{black} Robots that operate in space are very much limited due to the unique challenges encountered like communication latency, lack of power sources and extreme safety requirements. Current robots operating in space are either controlled from ground stations or tele-operated. As a result, there is large scale research going on to make space robots more autonomous.  

One of the critical issues that require immediate attention is the ever increasing space junk, especially in the past decade as it poses a huge threat to the functioning spacecraft (satellites, International Space Station etc.). There are more than half a million debris in Lower Earth Orbit (LEO)\cite{spacedebris}  and it  is  estimated  that  the  space  environment  can be stabilised when on the order of 5–10 objects are removed from LEO per year \cite{ESA}. 
\begin{figure}[h]
    \centering
    \includegraphics[width=1\columnwidth]{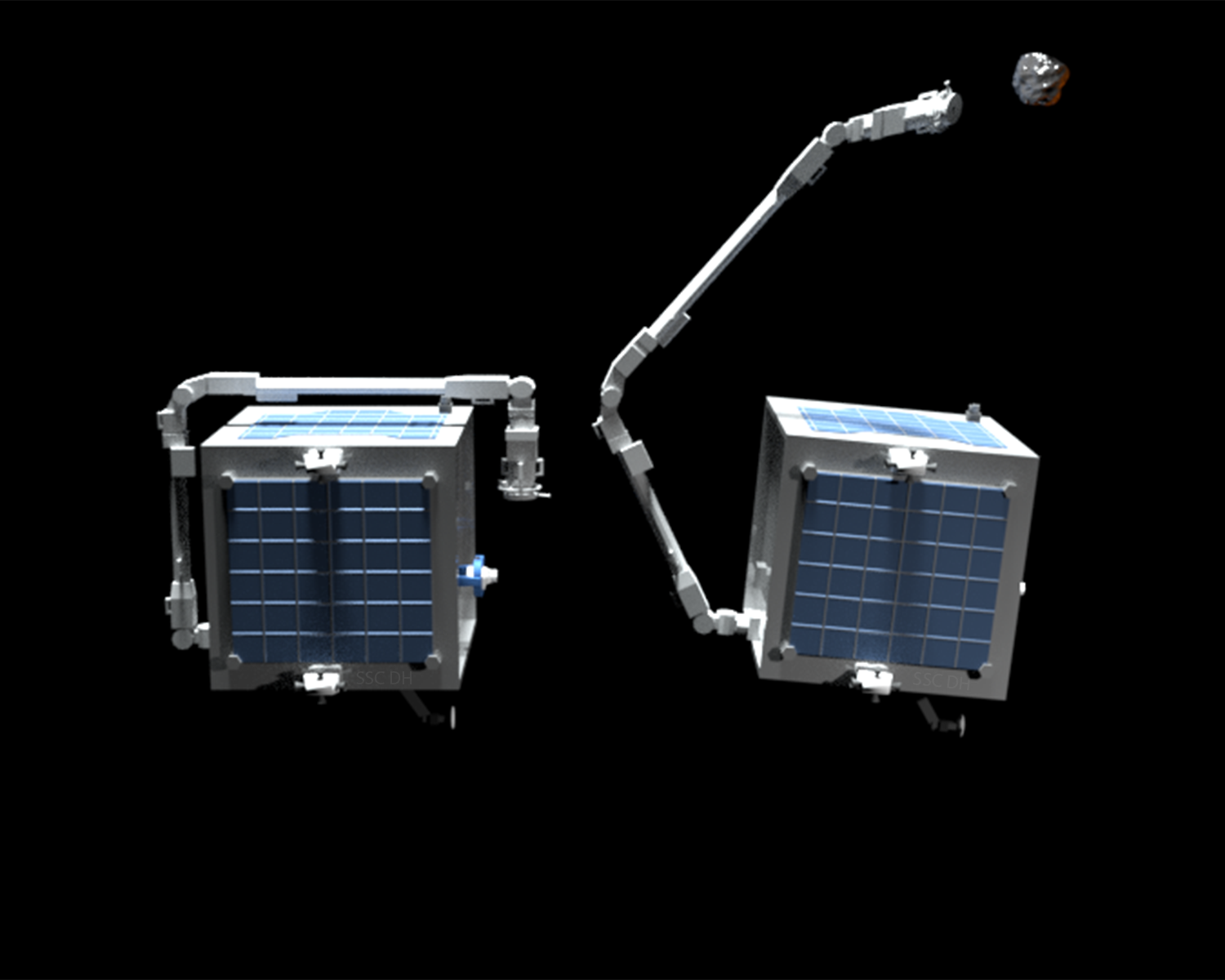}
    \caption{Left: Home position of the Future Debris Removal Orbital Manipulator (SDROM) (a 7-DoF redundant robot arm attached to the spacecraft); Right: Position of the Future Debris Removal Orbital Manipulator (SDROM) with Manipulator capturing an orbital debris.}
    \label{fig:conceptual}
\end{figure}
Although several methods for space debris removal has been proposed like harpoons, nets, tentacles \cite{spacedebris_methods}, using robotic arms to capture still remain the preferred choice as it can be extended to various other application areas like  on-orbit servicing \& assembly and autonomous rendezvous \& docking. 

Spacecraft require flying at a nominal attitude to charge battery, communicate with the ground station and determine its attitude and position. However orbital environment being micro-gravitational poses a difficult challenge since the spacecraft bus to which the robot-arm is attached is floating and any motion of the robot-arm would induce an attitudinal disturbance to the spacecraft. For free-flying spacecraft, the attitude determination and control system (ADCS) continuously compensate for the disturbances from the operation of the manipulator to maintain the nominal attitude of the spacecraft and hence a lot of energy is consumed.

Free-floating is a conceptual operating state of the spacecraft (when ADCS is switched off) installed with robotic arm. This type of spacecraft leave the attitude uncontrolled during the operation of the robotic arm. However, leaving the attitude of the spacecraft tumbled is unsafe and not ideal for the power system and the sensors used for determining its attitude. For both cases, we want the trajectory planner of the robotic arm to consider two important aspects, viz.,

\begin{enumerate}
    \item minimal attitudinal disturbances of the spacecraft bus due to manipulator operation
    \item computationally inexpensive for minimal power consumption. 
\end{enumerate}

In this work, we demonstrate how programming by demonstrations or imitation learning \cite{ashbabu, paraschos2018using}  can be used to plan trajectory of a 7-DoF robot arm attached to small spacecraft. The learned trajectories are efficiently encoded as a probabilistic distribution (PD) from which we can sample out trajectories for reproduction. This method is computationally efficient and is capable of minimizing the attitude disturbances (as shown in \cref{subsec:attittude_optim}). 

Optimal control methods \cite{rybus2016trajectory, rybus2017control, camacho2013model} are well developed but it often get stuck at local minima due to  poor initial guess and if successful, produces only a single trajectory . However, modeling as a PD captures the mean as well as the variance of the trajectories. The variance information could be used for sampling initial guess values form the PD for optimization based local planners to perturb the trajectory to avoid obstacles. It is known that the quality of initial guess determines the computational load and avoidance of local minima \cite{ashbabu}. 

Future space robots is expected to have human arm like dexterity. Hence it is proposed to use a 7-DoF redundant robot arm. This has the added advantage that the planning and control can still be carried out effectively even if a joint encoder or sensor fails.

%% file: src/related_work.tex

The analysis of the kinematic and dynamic of spacecraft with manipulator is well established.
%
%
Exploiting the non-holonomic behavior of the orbital manipulator system for spacecraft attitude and end-effector trajectory control have been studied extensively. It usually involves joint space techniques to control both the motion of the arm and sometimes the spacecraft attitude \cite{Yoshida, Hirano2018}. Early research used mapping methods to correlate the end-effector position with the induced disturbances on the spacecraft to minimize the attitude disturbances \cite{Torres1992, Vafa1993}. However, the mapping methods are computationally inefficient and furthermore, higher DoF manipulators will significantly increase the mapping difficulty and are challenging to find optimised paths.

The work by Nenchev et.al. \cite{Nenchev1999} proves that for certain manipulator motions, no reaction forces are induced on the spacecraft. As mentioned in their work, such solutions exists only for some special cases where integrability of the reaction null space velocity exists.  
%
%
This work then inspired many following research to exploit and optimise the control method for spacecraft with manipulator \cite{Dimitrov2006, Piersigilli2010,Nguyen-Huynh2013}.

More recently, researchers have attempted to solve the problem of trajectory planning by minimizing a cost functional which satisfies certain criteria. For example \cite{rybus2016trajectory, seweryn2008optimization} minimizes the power consumption. Non-linear Model Predictive Control (NMPC) have been used for control of free-floating spacecraft \cite{rybus2017control} but it remains to be seen how such heavy computations can be carried out by an on-board spacecraft computer. The other focus is post-panning impedance control of the orbital manipulator to free-motion targets. These researches aim to solve the kinodynamics in order to finely control the impact force for safe and accurate manipulation in the micro-gravity environment \cite{Papadopoulos}.



\subsection{Contributions}
The main contributions of this work are 
\begin{enumerate}
    \item Imitation learning based trajectory planning: \begin{itemize}
        \item First the trajectories are learned from demonstrations and encoded as a probabilistic distribution (PD). Planning to an unseen target only requires  sampling and conditioning of the PD. This avoids computationally expensive optimization methods (which usually have a cost function to minimize) to run on on-board space computer.
    \end{itemize} 
    \item Minimize attitude disturbances during capture: 
    \begin{itemize}
        \item Sampling from a PD for our redundant manipulator arm can produce infinite possible trajectories theoretically. Attitude disturbances for each trajectory can be easily computed and is possible to choose the trajectory with the least disturbance.
    \end{itemize}
\end{enumerate}

This paper is organized as follows. \Cref{sec:dynamic_formulation} gives briefly  kinematic and dynamic formulations of the spacecraft manipulator system. In \cref{sec:trajectory_learning}, we discuss the method used for generating trajectory data for learning. \Cref{sec:traj_encoding} provides the equations by which trajectories can be compactly encoded as a probabilistic distribution which can be used further for reproduction to unseen situations. \Cref{sec:simulation} presents the simulation results and \cref{sec:conclusion} gives the conclusions and future directions.


%% file: src/dynamic_formulation.tex
{\color{black} 
The kinematic and dynamic formulation of free-floating spacecraft have been studied previously \cite{umetani1989resolved, wilde2018equations, nanos2017dynamics}. Here we give a abridged version of the same for completeness. This formulation makes it easier to compute various matrices especially the coriolis and centrifugal which requires symbolic differentiation of the mass matrix. The whole formulation is carried out using Python's symbolic library called 'sympy' \cite{sympy} which supports 'C' code generation as well for faster execution. \\

\subsection{Nomenclature}
\begin{itemize}
    \item $m_i$ : mass of the $i^{th}$ link, the first being the spacecraft
    \item $r_i$ : position vector of the centre of mass of $i^{th}$ link with respect to the inertial co-ordinate system
    \item $\dot{r}_i$ : linear velocity of the centre of mass of $i^{th}$ link with respect to the inertial co-ordinate system
    \item $I_i$ : moment of inertia of $i^{th}$ link with respect to the inertial co-ordinate system
    \item $\omega_i$ : angular velocity of the $i^{th}$ link with respect to the inertial co-ordinate system
    \item $a_i$ : vector pointing from the joint $i$ to the centre of mass of link $i$
    \item $b_i$ : vector pointing from the centre of mass of link $i$ to joint $i$ + 1
    \item $l_i$ : length of $i^{th}$ link
    \item $\phi_s$ : vector of attitude angles (yaw, pitch and roll) of the spacecraft
    \item $\phi_m$ : vector of manipulator joint angles
\end{itemize}

\subsection{Assumptions}
\begin{enumerate}
    \item Momenta is conserved and is zero at the beginning
    \item Gravity is negligible
    \item The Centre of Mass of the system coincides with the origin of the inertial co-ordinate system
    \item The motion planning is carried out when the satellite-manipulator system at a safe state and is sufficiently close to the target.
\end{enumerate}
The mass centre of the spacecraft-manipulator arm can be described as
\begin{align}
    \sum_{i=0}^n m_i r_i = 0
    \label{eqn:com}
\end{align}
The linear and angular momentum conservation equations become
\begin{align}
    \sum_{i=0}^n m_i \dot{r_i} = 0 \\
    \sum_{i=0}^n I_i \omega_i = 0
\end{align}
\begin{figure}[h]
    \centering
    \title{Demonstration trajectories}
    \includegraphics[width=0.9\columnwidth]{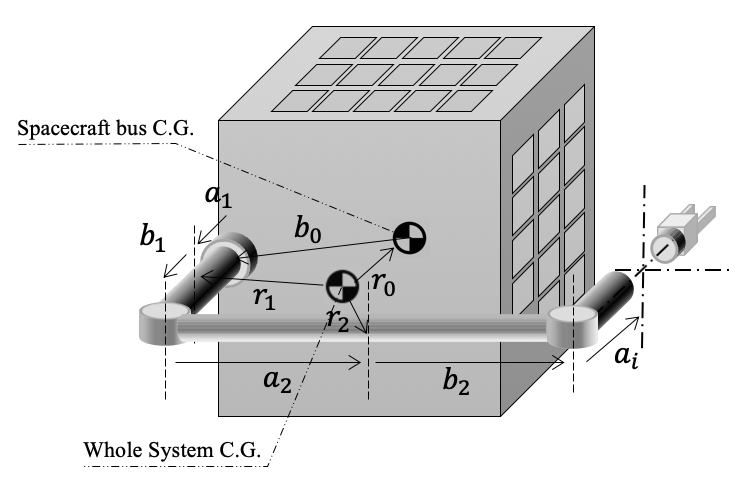}
    \caption{Schematic diagram of a spacecraft-manipulator arm}
    \label{fig:schematic}
\end{figure}
From \cref{fig:schematic}, the geometrical relationship between the  various vectors can be written as
\begin{align}
    r_{i} = r_{i-1} + a_{i} + b_{i-1}
    \label{eqn:geom_rel}
\end{align}
\Cref{eqn:com} and \cref{eqn:geom_rel} can be solved simultaneously to obtain
the centre of mass of the spacecraft and can be expressed as in \cref{eqn:r_s}
\begin{align}
    r_s = r_0 &= -\sum_{i=0}^{n-1} K_{ij} (b_i + a_{i+1}) \label{eqn:r_s}\\
    K_{ij} &= 1 - \sum_{j=0}^i \frac{m_j}{W} \nonumber \\
    v_s &= \frac{d}{dt}r_s \label{eqn:spacecrafLinVelocity}
\end{align}
where $W$ is the total mass of the system, $r_s~  \mathrm{and} ~v_s$ are the position vector and linear velocity of the spacecraft with all vectors expressed with respect to the inertial co-ordinate system. The position vector and velocity of the rest of the links can be found using the recursive relation given by \cref{eqn:geom_rel}. The differential kinematics of the satellite-manipulator arm system gives the jacobian matrix of the system which consists of the manipulator part ($J_m$) and the satellite part ($J_s$). Thus the end-effector velocity, $v_{eef}$, and the momentum conservation can be expressed as in \cite{umetani1989resolved}

\begin{subequations}
\label{eqn:ang_mntm_conser}
\begin{align}
    v_{eef} &= J_s \dot{\phi}_s + J_m \dot{\phi}_m \\
    0 &= I_s \dot{\phi}_s + I_m \dot{\phi}_m \label{eqn:ang_mntm}
\end{align}
\end{subequations}
From \cref{eqn:ang_mntm_conser}, the end-effector velocity can be solved as a function of the manipulator joint rates and generalized jacobian, $J^*$ given by $(J_m - J_s I_s^{-1} I_m)$
\begin{align}
    v_{eef} &= (J_m - J_s I_s^{-1} I_m) \dot{\phi}_m \nonumber \\
    &= J^* \dot{\phi}_m
    \label{eqn:gen_jacob}
\end{align}
where $I_s$ and $I_m$ are respectively the satellite and manipulator inertia matrices expressed in inertial co-ordinate system \cite{umetani1989resolved}.

The Kinetic energy, T,  can then be expressed as 
\begin{align}
    T = \sum_{i=0}^n m_i (v_i ~.~ v_i) = \frac{1}{2} \dot{\phi}^T M(\phi)~ \dot{\phi}
\end{align}
where $M(\phi)$ is the mass matrix and $\phi = [\phi_s^T~ \phi_m^T]^T$.
The centripetal and coriolis vector, C, is given by
\begin{align}
    C(\phi, \dot{\phi}) = \dot{M}(\phi)~ \dot{\phi} - \frac 12 
    \begin{bmatrix}
    	\dot{\phi}^T \frac{\partial M(\phi)}{\partial \phi_1} \dot{\phi} \\ 
    	\dot{\phi}^T \frac{\partial M(\phi)}{\partial \phi_2} \dot{\phi} \\ 
    	. \\ 
    	. \\ 
    	\dot{\phi}^T \frac{\partial M(\phi)}{\partial \phi_n} \dot{\phi}
    \end{bmatrix} 
\end{align}
The equation of motion of the free-floating spacecraft manipulator system can be written as
\begin{align}
    M(\phi) \ddot{\phi} + C(\phi, \dot{\phi}) = \begin{bmatrix}
    0 \\
    \tau
    \end{bmatrix}
\end{align}
where $\tau$ is the control torque to be applied at the manipulator joints.
}

%% file: src/data_generation_mpc.tex
{\color{black}
The method introduced here requires data samples for trajectory learning. A trajectory, $\zeta$, is a mapping of all the robot configuration ($x$) from start to goal with time.  Mathematically it can be represented as
$\zeta: [0, 1] \longrightarrow x$ where $x \in \mathbb{R}^d $ and $d$ corresponds to the number of joints with $\zeta(0)$ and $\zeta(1)$ being the start and goal configurations respectively. These trajectories could be generated by a human expert by demonstrations \cite{zhu2018robot, havoutis2019learning}. As real hardware orbital simulation of micro-gravity environment being extremely expensive, we demonstrate the concept by generating trajectories using an optimal control algorithm \cite{kirk2004optimal}. We make use of the redundancy of the chosen 7-DoF manipulator arm to generate several trajectories which starts at the home position (\cref{fig:conceptual}) and go to a particular goal state given by the vision system. Once enough trajectories are generated, the goal state is changed and the process is repeated until the entire workspace is  covered.

The cost function for trajectory generation is given as
\begin{align}
    J = x_T^T P_T x_T + \int_{t_0}^{T-1} x_t^T Q_t x_t + u_t^T R_t u_t ~dt
    \label{eqn:opt_control}
\end{align}
subject to the constraints 
\begin{align*}
    \dot{x}_t &= A_t x_t + B_t u_t \\
    x_t(0) &= x_0
\end{align*}
where $x_t$ represents the state (position \& velocity in joint space) of the manipulator joints at time $t$. For a space-manipulator, once the manipulator states have been found out, the satellite states can be determined from \cref{eqn:ang_mntm_conser} and integration. Here $Q_t$ and $R_t$ are respectively the time varying state and control cost matrices, $P_T$ is the stabilizing matrix obtained by the solution of algebraic Ricatti equation at every time instant \cite{bemporad2002explicit}, $t_0, T$ are the initial and final time respectively. The trajectory data samples in simulation are obtained by the following methods.
\begin{enumerate}
    \item varying the cost matrices thus encouraging certain joint motions and discouraging certain other joint motions.
    \item introducing artificial obstacles (elastic bands \cite{quinlan1993elastic}) between the start and goal point so as to force the redundant robot to follow a different trajectory to the same goal point
    \item introduction of noise into the constraint equation of the system given by \cref{eqn:opt_control}.
\end{enumerate}  

}

%% file: src/trajectory_encoding.tex
{\color{black}The generated trajectory data samples need to be represented in an efficient manner for future planning and control. It has to be mentioned that the trajectory planning is carried out when the spacecraft-robot arm is sufficiently close to the target and is safe to operate. Here we demonstrate the core idea of this work by representing the generated trajectories as a probabilistic distribution. We find that Gaussian distributions fit all the essential criteria for efficiently representing trajectories as it depends only on two parameters i.e. mean and covariance. For reproduction of trajectories to unseen situations, we use the conditioning property of the Gaussian as explained in \cref{subsec:conditioning}.

\subsection{Gaussian trajectory encoding}
\label{subsec:lfd}
The encoding of the trajectories can be expressed as a linear basis function model as in \cref{eqn:lin_basis_fn_model}  where $\boldsymbol{\psi_t}$ = $[\psi_t ~ \dot{\psi_t}]^T$ is the basis function (see APPENDIX \ref{app:basis_function}), $w$ a parameter vector, plus some error $\epsilon$. Such a representation reduces the number of parameters and facilitates learning. Assuming trajectories to be independent and identically distributed, the probability of observing a trajectory, $\zeta$, given the parameter vector $w$ can be written as in \cref{eqn:gau_traj_encode} \cite{paraschos2018using}
\begin{align}
    x_t &= \begin{bmatrix}
    q_t \\ 
    \dot{q}_t
\end{bmatrix} = \begin{bmatrix}
    \psi_t \\ 
    \dot{\psi}_t
\end{bmatrix}  w + \epsilon_x \label{eqn:lin_basis_fn_model} \\
p(\zeta|w)&=  \prod_{t} \mathcal{N}(x_t|\pmb{\psi_t} w, \Sigma_x ) 
\label{eqn:gau_traj_encode}
\end{align}
where  $\epsilon_x \sim \mathcal{N} (0,~ \Sigma_x)$ represents the zero-mean Gaussian noise associated with each observation and $x_t$ is the state. There are several possible choices for the basis function. Here we use a Gaussian radial basis function representing stroke based movements which is ideal for motion planning \cite{paraschos2018using}. 
The parameter vector $w$ is modeled as another Gaussian distribution with parameter $\theta~ = \{\mu_w, \Sigma_w$\} to capture the variance of the trajectories. (Here $\mu_w$ and $\Sigma_w$ are respectively the mean and variance of the Gaussian).
Using the linear transformation property of the Gaussian distribution (see APPENDIX \ref{app:lin_trans_gaussian}), the state can be represented as 
\begin{align}
p(x_t; \theta) &= \int \mathcal{N}(x_t| \pmb{\psi_t} w, \Sigma_x) \mathcal{N}(w | \mu_w, \Sigma_w) dw \nonumber \\ &= \mathcal{N}(x_t|\mu_w, \pmb{\psi_t} \Sigma_w \pmb{\psi_t}^T + \Sigma_x )
\end{align}
For the generated trajectory samples, the parameter vector $w$ can be estimated as a ridge regression and is given in \cref{eq:ridge}
\begin{align}
    w_i = (\Psi^T\Psi + \lambda I)^{-1}~ \Psi^T X_i
    \label{eq:ridge}
\end{align}
where $X_i$ is a 1-D concatenated vector (see APPENDIX \ref{app:learning scheme}) of all joint values  during all time steps from the $i^{th}$ trajectory sample and $\Psi$ is a block diagonal matrix with each block diagonal being $\pmb{\psi_t}$.
The mean and variance of the parameter vector, $w$, are estimated as in \cref{eq:mean&varWeight}

\begin{align}
    \mu_w &= \frac{1}{N} \sum_{i=1}^N w_i \nonumber\\
    \Sigma_w &= \frac{1}{N}\sum_{i=1}^N(w_i - \mu_w)(w_i - \mu_w)^T
    \label{eq:mean&varWeight}
\end{align}
where $N$ is the number of demonstrations.

All the above computations could be carried out once the spacecraft manipulator design is complete and the whole trajectory planning problem could then be stated as taking the robot from $\zeta(0)$ (home position) to $\zeta(1)$ which is the pose of the target estimated by the vision system.

\subsection{Trajectory planning to unseen situations}
\label{subsec:conditioning}
The data generation described in \cref{sec:trajectory_learning} needs several trajectory samples to accurately represent the workspace of the spacecraft-manipulator. However, workspace can have infinite possible locations of the target theoretically and it is impossible to do data generation for all possible goal poses. The Gaussian distribution introduced above can solve this problem by using the conditional distribution property.
A probabilistic trajectory distribution can be conditioned to follow not only the desired start and goal state but also the via-points~\cite{paraschos2018using}.
For example, if our trajectory has to pass through a desired state $x^*_t$ the new mean and variance of the conditioned trajectory will be  
\begin{align}
    \mu_w^{[new]} &= \mu_w + L(x^*_t - \pmb{\psi_t}^T\mu_w) \nonumber\\
    \Sigma_w^{[new]} &= \Sigma_w - L\pmb{\psi_t}^T\Sigma_w
    \label{eq:conditioning}
\end{align}
where $L$ is 
\begin{align}
    L = \Sigma_w\pmb{\psi_t}(\Sigma^*_x + \pmb{\psi_t}^T\Sigma_w\pmb{\psi_t})^{-1}    
\end{align}
and $\Sigma^*_x$ is the desired accuracy to which the the state ($x^*_t$) is to be reached
\subsection{Cost of a trajectory}
\label{subsec:attittude_optim}
The cost of a trajectory is a scalar which estimates how much the attitude of the spacecraft changes when the robot arm follows a particular trajectory. It is defined as follows.
%
%
\begin{align}
    Q =  c^2 ~\Sigma ~ \dot{\phi}_s^T ~ \dot{\phi}_s + \Sigma ~ v_s^T v_s
\end{align}
where $\dot{\phi}_s$ (from \cref{eqn:ang_mntm}) and $v_s$ (from \cref{eqn:spacecrafLinVelocity}) are respectively the angular and linear velocities of the spacecraft calculated at each discrete time step from initial to final pose, $c$ is a angular to linear conversion coefficient which allows to combine an angular value with a linear value and $\Sigma$ is the summation symbol. The minimum cost corresponds to the trajectory having minimal disturbances.
\subsection{Algorithm}
The algorithm can be summarised as given below

\begin{itemize}
    \item \textbf{Given:}
        \begin{itemize}
        \item[--] Start, $\zeta(0)$
            \item[--] Trajectory data from the optimal control (joint parameters)
            \item[--] DH table of the robot
        \end{itemize}
    \item \textbf{Precompute:}
        \begin{itemize}
    	    \item[--] Compute the basis function, $\Psi$
    	    \item[--] Compute the mean, $\mu_w$ and covariance, $\Sigma_w$ of the parameter vector from trajectory data.
        \end{itemize}
    \item \textbf{Repeat for multiple debris removal:}
        \begin{itemize}
        	\item[--] Estimate target pose from vision system, $\zeta(1)$
        	\item[--]
        	Compute $\mu_w^{new}$ and $\Sigma_w^{new}$ 
        	\item[--] Sample trajectories (say 100) from the conditioned distribution using $\mu_w^{new}$ and $\Sigma_w^{new}$
        	\item[--] For each trajectory, find the cost
        	\item[--]Select the trajectory having minimum cost
        \end{itemize}  
\end{itemize}
}

%% file: src/simulation.tex
The Denavit-Hartenberg~\cite{hartenberg1955kinematic} parameters of the robot arm is shown in \Cref{tab:DH}
and the parameters used for the simulation are shown in \Cref{tab:sim_par}.
\begin{table}[h]
\caption{DH parameters of the robot arm}
\label{tab:DH}
\centering
\renewcommand{\arraystretch}{2.5}
\begin{tabular}{|c|c|c|c|c|}
\hline
\textbf{Joint} & \textbf{$\alpha$ (rad)} & \textbf{$a$ (m)} & \textbf{$d$ (m)} & \textbf{$\theta$ (rad)} \\ \hline
1              & -$\dfrac{\pi}{2}$        & 0.0              & 0.5              & $\theta_1$              \\ \hline
2              & $\dfrac{\pi}{2}$         & 0.0              & 0.0              & $\theta_2$              \\ \hline
3              & $\dfrac{\pi}{2}$         & 0.9              & 0.0              & $\theta_3$              \\ \hline
4              & -$\dfrac{\pi}{2}$        & 0.9              & 0.0              & $\theta_4$              \\ \hline
5              & $\dfrac{\pi}{2}$         & 0.8              & 0.0              & $\theta_5$              \\ \hline
6              & -$\dfrac{\pi}{2}$        & 0.8              & 0.0              & $\theta_6$              \\ \hline
7              & $\dfrac{\pi}{2}$         & 0.0              & 0.8              & $\theta_7$              \\ \hline
\end{tabular}
\end{table}
\begin{table}[h]
\caption{Simulation Parameters}
\label{tab:sim_par}
\centering
\renewcommand{\arraystretch}{1.}
\begin{tabular}{|l|l|l|l|l|l|l|l|l|}
 \hline
    & \begin{tabular}[c]{@{}l@{}}Satellite\\ L 0\end{tabular} & L 1 & L 2 & L 3 & L 4 & L 5 & L 6 & L 7\\ \hline
Mass (kg) & 200.0  & 20.0 & 30.0 & 30.0 & 20.0 & 20.0 & 20.0 & 20.0   \\ \hline
$I_x$     & 1400.0  & 0.10    & 0.25   & 0.25 & 0.25   & 0.25 & 0.25   & 0.25   \\ \hline
$I_y$     & 1400.0  & 0.10   & 25.0   & 25.0 & 25.0   & 25.0 & 25.0   & 25.0  \\ \hline
$I_z$     & 2040.0  & 0.10   & 25.0   & 25.0  & 25.0   & 25.0 & 25.0   & 25.0 \\ \hline
\end{tabular}
\end{table}

\begin{figure}[h]
    \centering
    \includegraphics[width=0.9\columnwidth, height=0.35\textwidth]{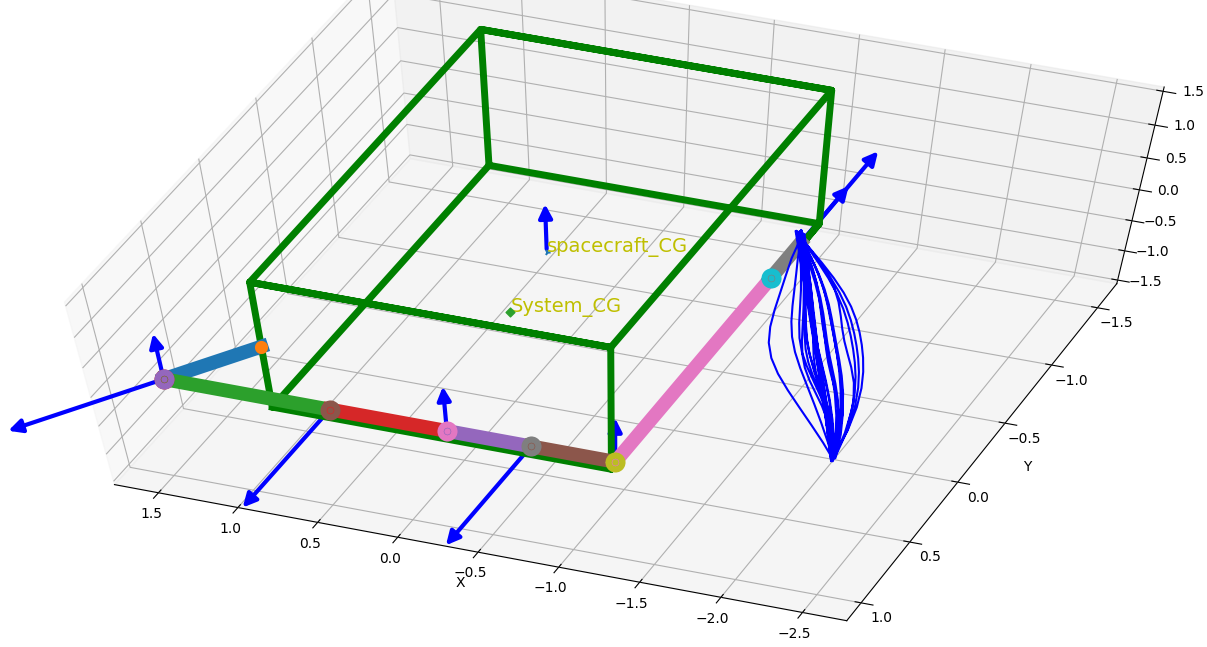}
    \caption{Conditioned trajectories of the end-effector sampled from the Probabilistic Distribution}
    \label{fig:end_effector_traj}
\end{figure}
\begin{figure}[h]
    \centering
    \includegraphics[width=0.9\columnwidth, height=0.35\textwidth]{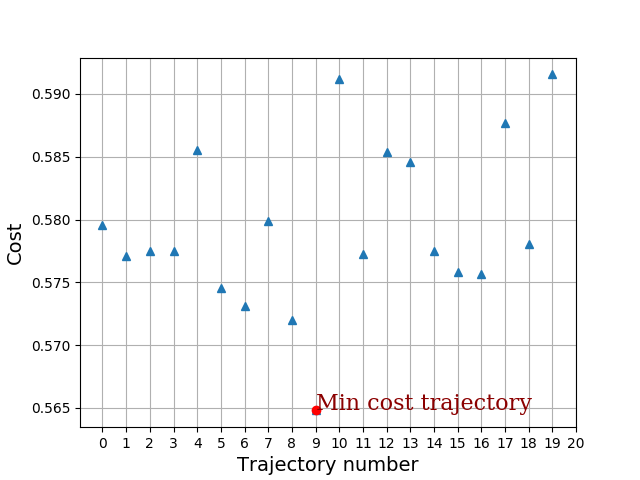}
    \caption{Cost associated with each of the sampled trajectory}
    \label{fig:traj_cost}
\end{figure}

\begin{figure}[h]
  \centering
  \subfigure[Spacecraft rotation]{\includegraphics[width=0.46\columnwidth, height=0.44\columnwidth]{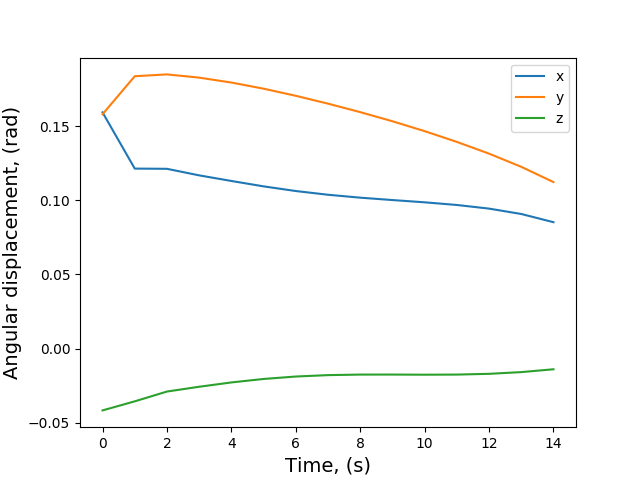}}\hspace{0.4cm}%
  \subfigure[Spacecraft translation]{\includegraphics[width=0.46\columnwidth, height=0.44\columnwidth]{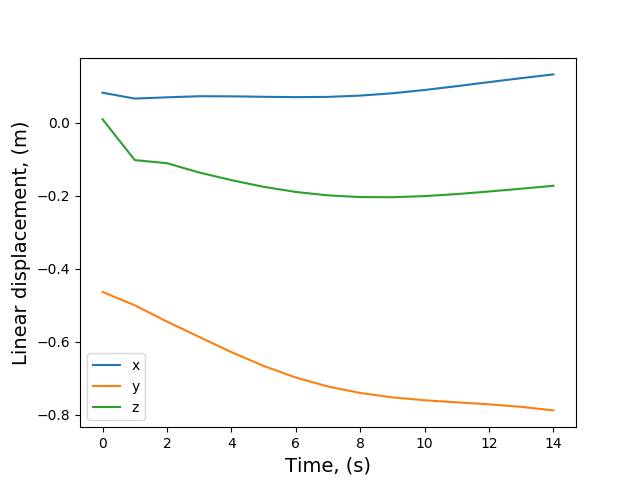}}\\
  \caption{Pose variation of the spacecraft CG during trajectory tracking}
  \label{fig:induced_spacecraft_motion}
\end{figure}

\cref{fig:end_effector_traj} shows the home position of the robot arm attached to the spacecraft corresponding to the joint values, $(0.0,~ 5\dfrac{\pi}{4},~ 0.0,~ 0.0,~  \dfrac{\pi}{2},~ -\dfrac{\pi}{2},~ 0.0)$. Twenty trajectories in joint space are sampled out from the conditioned distribution. For the twenty joint space trajectories, the corresponding end-effector trajectories in task space are found out and are shown in \cref{fig:end_effector_traj}. \Cref{fig:traj_cost} shows the cost of each of the trajectories and the trajectory which has the minimum cost. For this trajectory, the induced motion on the spacecraft is shown in \cref{fig:induced_spacecraft_motion}

%% file: src/conclusion.tex
To the authors' knowledge, this is the first time imitation learning is used in space robot arm trajectory planning for free floating spacecraft. This work addresses the issue of minimizing attitude disturbance spacecraft bus when the arm reaches out to capture a debris. The learning is carried out offline and is computationally very efficient for finding new trajectories after deployment. 

The trajectory learning algorithm presented in this paper will be potentially tested on a Future Space Debris Removal Orbital Manipulator (SDROM) which has a similar micro-satellite spacecraft bus as RemoveDEBRIS \cite{Forshaw2017} but with a 7-DoF redundant robot arm attached, as shown in \cref{fig:conceptual}. This is the next step towards space autonomy for on-orbit operations that will be demonstrated by a potential mission concept that goes beyond RemoveDEBRIS spacecraft. The overall mission objectives would be to execute pose estimation, trajectory and motion planning of the robotic arm, and capture a sample debris in order.

One of the possible extensions is to use the variance information to find out obstacle free trajectories when the operating space is cluttered. Another way forward is devising a control strategy from the learned distribution and simulate in a physics engine

%% file: src/appendices.tex
\begin{appendices}
\section{Basis function}
\label{app:basis_function}
\begin{figure}[h]
    \centering
    \includegraphics[width=0.9\columnwidth, height=0.35\textwidth]{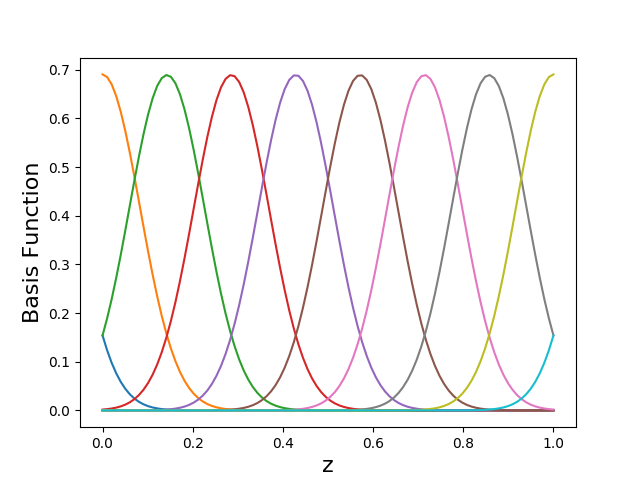}
    \caption{Radial basis function}
    \label{fig:basisfn}
\end{figure}

The basis function used in this work is given by

\begin{align*}
    \psi_i(z) = \rm{exp}^{\Big(-\dfrac{(z - c_i)^2}{h^2}\Big)}
\end{align*}
where $z_t$ is the phase, $c_i$ is the center and $h$ is the bandwidth factor. Interested readers can refer to \cite{paraschos2018using} for details. \cref{fig:basisfn} gives a plot of ten basis function centered at [-0.142,  0.,  0.142, 0.285, 0.428, 0.571, 0.714, 0.857, 1.,  1.142]
\section{Linear transformation property of Gaussian distribution}
\label{app:lin_trans_gaussian}

If a random variable $x$ is normally distributed ($\mathcal{N}(\mu,\Sigma)$), the linear transformation $Ax + c$ follows the distribution
\begin{align*}
Ax+ c \sim \mathcal{N}(A\mu + c, A \Sigma A^T)
\end{align*}

\section{Explanation of the learning scheme}
\label{app:learning scheme}
To illustrate the learning process, a simple example of 2-DoF planar robot is considered here. Let $\theta_1$ and $\theta_2$ represent the joint angles. The complete trajectory for a single demonstration, $\zeta$,  concatenated as a 1-D vector is represented as
\begin{align*}
    \lambda_1 &= [\theta_{1_{t1}}, \theta_{1_{t2}}, ..., \theta_{1_{tn}}]\\
    \lambda_2 &= [ \theta_{2_{t1}}, \theta_{2_{t2}}, ..., \theta_{2_{tn}}] \\
    \zeta &= [\lambda_1 \lambda_2, \dot{\lambda}_1, \dot{\lambda}_2]^T_{tn \times 2nDoF}
\end{align*}
Here $t_i$ are the time points, $tn$ and $nDoF$ are the total time points and number of degrees of freedom respectively. Let the number of basis functions be $nBf$. The matrices $\psi_t$ and $\dot{\psi}_t$ are of dimension $2nDoF \times (nBf \times 2nDoF)$ each and the matrix $\Psi$ is of dimension $(tn \times 2nDoF) \times (nBf \times 2nDoF)$. The learning parameter, $w_i$ for one demonstration is then calculated by formulating as a regression problem and reducing the loss, $(\zeta - \Psi w)^2$ 
\end{appendices}